\documentclass[conference, a4paper]{IEEEtran}
\IEEEoverridecommandlockouts
\usepackage{cite}
\usepackage{amsmath,amssymb,amsfonts}
\usepackage{algorithmic}
\usepackage{graphicx}
\usepackage{textcomp}
\usepackage{xcolor}

\IEEEoverridecommandlockouts\IEEEpubid{\makebox[\columnwidth]{ 979-8-3503-5229-0/24/\$31.00 ©2024 IEEE  \hfill} \hspace{\columnsep}\makebox[\columnwidth]{ }}

\begin{document}

\title{Filling in the Blanks: Applying Data Imputation in incomplete Water Metering Data
\thanks{
This work was partially supported by the European Union under the Italian National Recovery and Resilience Plan (NRRP) of NextGenerationEU, partnership on “Telecommunications of the Future” (PE00000001 - program RESTART - CUP E83C22004640001) in the WITS and SPRINT focused project.
This work was partially supported by the project SERICS (PE00000014 - CUP D33C22001300002) under the NRRP MUR program funded by the EU-NGEU. 
This work is supported by the European Union’s Horizon Program under the Agile and Cognitive Cloud edge Continuum management (AC3) project (Grant No. 101093129).}
}


\author{
\IEEEauthorblockN{Dimitrios Amaxilatis}
\IEEEauthorblockA{
\textit{Spark Works Ltd.}\\
Galway, Ireland \\
d.amaxilatis@sparkworks.net\\
ORCiD:0000-0001-9938-6211}
\and
\IEEEauthorblockN{Themistoklis Sarantakos}
\IEEEauthorblockA{
\textit{Spark Works Ltd.}\\
Galway, Ireland \\
tsaradakos@sparkworks.net\\
ORCiD:0000-0002-7517-6997}
\and
\IEEEauthorblockN{Ioannis Chatzigiannakis}
\IEEEauthorblockA{
\textit{Sapienza University of Rome}\\
Rome, Italy \\
ichatz@diag.uniroma1.it\\
ORCiD:0000-0001-8955-9270}
\and
\IEEEauthorblockN{Georgios Mylonas}
\IEEEauthorblockA{
\textit{Industrial Systems Institute,}\\
\textit{Athena Research Center}\\
Patras, Greece \\
ORCiD:0000-0003-2128-720X}
}

\maketitle

\begin{abstract}
In this work, we explore the application of recent data imputation techniques to enhance monitoring and management of water distribution networks using smart water meters, based on data derived from a real-world IoT water grid monitoring deployment.
Despite the detailed data produced by such meters, data gaps due to technical issues can significantly impact operational decisions and efficiency. 
Our results, by comparing various imputation methods, such as k-Nearest Neighbors, MissForest, Transformers, and Recurrent Neural Networks, indicate that effective data imputation can substantially enhance the quality of the insights derived from water consumption data as we study their effect on accuracy and reliability of water metering data to provide solutions in applications like leak detection and predictive maintenance scheduling.
\end{abstract}

\begin{IEEEkeywords}
Data imputation, Smart meters, Smart Water Grid, Data analysis, Smart City, IoT, Machine Learning
\end{IEEEkeywords}

\section{Introduction}

In the era of smart cities and advanced utility management, the monitoring of water grids has become increasingly pivotal to ensuring efficient distribution, sustainability, and infrastructure reliability. 
Smart water meters, integral components of this intelligent monitoring framework, offer a granular view of water usage, potential leaks, and system performance. 
However, despite their sophistication, the occurrence of missing data due to various factors—ranging from technical malfunctions to data transmission errors— remains an open challenge that undermines the integrity and actionable insights that can be derived from the datasets produced by such infrastructure.

Moreover, the significance of addressing missing data extends beyond mere data completeness. In the context of water grid monitoring, it impacts decision-making processes related to water management, leak detection, and predictive maintenance, all of which have profound implications for operational efficiency and environmental sustainability. Herein lies the critical role of data imputation methodologies designed to intelligently fill in missing data points and enhance the quality and usability of data collected by smart water meters.

This work explores the application of advanced data imputation techniques within the domain of water grid monitoring. 
By integrating these methods with data from smart water meters, we aim to not only address the gaps in data but also augment the accuracy and reliability of the insights gained from such data. 
Through a comprehensive review of existing imputation methods and their application to water grid consumption data, we endeavor to provide a robust framework for utility managers, policymakers, and researchers to optimize water resource management and infrastructure planning.
The data we use as a basis for our analysis come from water meters deployed in over 20 public buildings in Greece, collected over a period of more than 3 years between 2019 and 2024. 

Our methodology is structured into 3 distinct phases; we begin with data preprocessing, standardizing the collected data into a uniform format.
During this stage, we identify missing values and generate supplementary information that aids in estimating such gaps.
Furthermore, we partition the preprocessed data into training and validation datasets, which are crucial for subsequent analysis. The second phase involves training the models using the configured training dataset.
We conduct extensive evaluations across multiple models and configurations to identify the most effective approach. 
In the last phase, we deploy these models to estimate data.
Subsequently, we assess the estimations' accuracy and reliability to ensure the outputs' quality.
This approach ensures a thorough analysis and validation of the data imputation process.

In the following sections, we delve into the background of water grid monitoring, the challenges posed by missing data, and the pivotal role of smart water meters in contemporary water management. 
Subsequently, we will outline the methodology employed in applying data imputation techniques to water grid monitoring, followed by a detailed analysis of the results and their implications for the field. 
Through this investigation, we aim to contribute a meaningful discourse to the nexus of smart utilities and data science, providing actionable insights and methodologies to enhance the efficacy of water grid monitoring systems.

\section{Related Work}

The use of quality data in water metering systems is of high importance when building applications that provide solutions such as leakage detection or consumption profiling/prediction.
\cite{ALI2022100504} presents a comprehensive Internet of Things (IoT)-based solution for real-time water management and leakage detection in water distribution networks (WDN). 
\cite{iot1020026} explores an advanced IoT-based system for monitoring water leaks in irrigation systems using real-time data from a wireless sensor network integrating machine learning algorithms to process the collected data.
\cite{10076716} and \cite{10.1007/978-3-031-49361-4_5} present novel methods for enhancing the efficiency of smart water distribution networks using edge computing within a LoRaWAN architecture. It introduces combining an IoT network simulator and a WDN simulator data, as well as a classification platform, which operates at the network's edge to detect and manage water leakages efficiently.
Similarly, \cite{w14142187} explores the use of IoT and machine learning to enhance the monitoring and management of water consumption in residential settings. The approach incorporates various sensors within a household to collect real-time data on water usage, which is then analyzed using machine learning techniques to identify and predict consumption patterns.

In all the above cases, the quality of the collected data, and the completeness of the dataset used plays a crucial role in generating high-value predictions and estimations. 
The ability to have such a dataset is not always easy and guaranteed, due to the hassles of deploying and managing IoT systems and installations.
As a result, the ability to fill in the gaps of the missing data is of paramount importance.

The collection of studies on data imputation techniques encompasses a diverse array of innovative approaches tailored to diverse data scenarios. \cite{8502041} provides an insightful review of imputation methods ranging from simple deletion to advanced machine learning (ML) techniques, focusing specifically on water distribution systems (WDS).
It introduces a strategic ``top-down bottom-up'' method for selecting the most suitable imputation technique based on detailed data analysis and the characteristics of missing data, along with a comprehensive evaluation of each method's strengths and weaknesses.

\cite{9755143} proposes enhancing the Multivariate Imputation by Chained Equations (MICE) method by reshaping sensor data to strengthen the correlation between observed and missing data, thereby improving imputation accuracy. This approach is demonstrated through its application to water quality monitoring data from Vermont, showing a significant improvement in model accuracy with at least a 23\% increase in R² values.
\cite{8987530} examines the effectiveness of the K-Nearest Neighbors (K-NN) method across various scenarios with different missing data mechanisms. The study validates K-NN's robustness in maintaining high data integrity and closely matching the accuracy of complete datasets, underscoring its utility in diverse conditions.
\cite{stekhoven2012missforest} introduces MissForest, a non-parametric method that leverages the random forest approach for imputing missing values in datasets containing both continuous and categorical variables. This method capitalizes on the random forest’s ability to handle complex data structures and interactions without a predefined parametric form, demonstrating superior performance over other imputation methods, particularly in complex datasets.
\cite{DU2023119619}'s SAITS model employs a self-attention mechanism to enhance the imputation of missing values in multivariate time series data, showing superior performance through extensive experiments on real-world datasets. This method’s robustness is highlighted by its dynamic adjustment of weights in response to missing information.
\cite{CHEN2021126573} presents the TrAdaBoost-LSTM model, which combines LSTM neural networks with instance-based transfer learning to address large-scale consecutive missing data, particularly in water quality datasets. This model significantly improves imputation accuracy by leveraging existing complete datasets to enhance missing data prediction.
\cite{wu2022timesnet} introduces TimesNet, a novel model that addresses complex temporal variations by transforming 1D time series data into 2D tensors. 
This transformation allows for the effective application of 2D convolutional operations, enhancing the model's performance across tasks like forecasting, classification, and anomaly detection.

Concerning water quality monitoring, \cite{ZHANG202263} provided a review and comparison of various approaches to the handling of missing data in near real-time environmental monitoring systems. A missing data imputation system design and implementation is also presented, with several similarities to our work. The authors conclude that the size of the missing data and the method selected greatly affect data imputation performance, with large data gaps dealt with significantly better by neural network-based methods.

Lastly, \cite{che2018recurrent}'s GRU-D model incorporates missing data patterns directly into its architecture, thus enhancing the handling and prediction accuracy of multivariate time series data. This model is proven through real-world clinical datasets and synthetic data, showing superior performance compared to other methods and contributing significantly to the fields of health informatics and time series analysis.

Based on all the aforementioned solutions, we investigate how they can be used to create a specific profile for the consumption of each building or customer to more accurately estimate their missing data against a one-fits-all solution using individually trained ML models.
\section{Tethys Dataset}
The dataset we use for the work showcased in this paper comes from the deployment of our Edge-Computing–Ready Water Metering System for Smart Cities called \textbf{Tethys}.

\subsection{Tethys Testbed Setup}
Since its first implementation in 2019 at the campus of the Aristotle University of Thessaloniki, Greece, the Tethys~\cite{amaxilatis2020smart} system has been actively monitoring water consumption in 22 university buildings including teaching and research facilities as well as a university hospital.
The installation of the water metering infrastructure included setting up new electronic water meters and integrating existing ones with IoT-enabled smart meters from 2 manufacturers, Sensus and Kamstrup.
The Tethys deployment uses wM-Bus to collect data from the deployed smart meters, through intermediate LoRaWAN-to-wM-Bus bridge devices we call Mox nodes.
The active deployment is composed of 28 Mox nodes and over 50 water metering devices.
Regarding the reporting periods of each smart meter, data is transmitted at varying intervals — from every three minutes to hourly — determined by the end device hardware specifications.
The data generated have already been used in other work already published, like \cite{zecchini2021identifying} regarding the impact of the COVID-19 pandemic on water usage within each monitored building, the corresponding user behaviors, and differences in the consumption profiles before, during, and after the outbreak of the pandemic.
This data, allows us to apply and evaluate our solution on buildings with very different consumption profiles and operation settings, including low and high consumption periods, differences between weekdays and weekends, holidays, and even a period where visitors to the buildings were very limited during the COVID-19 pandemic.

\subsection{Dataset Description}
The dataset we are using to impute the missing data from our deployment consists of hourly aggregated data from the data received from the smart meter deployment.
We use hourly aggregates to remove inconsistencies from the different reporting rates of the metering infrastructure, as described above.
Table~\ref{tab:dataset} showcases the amount of data missing for each building, as well as the period that the infrastructure was operating, as the deployment was done incrementally, and certain devices were subsequently removed, due to reasons outside our scope.
As a result, the percentage of missing data for each building is calculated using the expected period of operation and not the total period between 2019 and 2023. 

\begin{table}[h]
    \centering
    \begin{tabular}{|c|c|c|c|c|}
        \hline
        Building    & Data  & Data      & Hourly        & \% of Hours   \\
        Id          & Since & Until     & Collected Data& Without Data  \\
        \hline
        0 & 2/2019 & 12/2023 & 33888 & 20 \% \\ 
        1 & 1/2019 & 12/2023 & 31779 & 26 \% \\ 
        2 & 5/2019 & 12/2023 & 30661 & 24 \% \\ 
        3 & 1/2019 & 12/2023 & 32338 & 25 \% \\ 
        4 & 5/2019 & 12/2023 & 31312 & 22 \% \\ 
        5 & 1/2019 & 12/2023 & 30501 & 29 \% \\ 
        6 & 1/2019 & 12/2023 & 33705 & 22 \% \\ 
        7 & 2/2019 & 3/2023 & 21483 & 39 \% \\ 
        8 & 1/2019 & 12/2023 & 9202 & 79 \% \\ 
        9 & 1/2019 & 12/2023 & 28863 & 33 \% \\ 
        10 & 2/2019 & 12/2023 & 32607 & 23 \% \\ 
        11 & 2/2019 & 12/2023 & 32507 & 23 \% \\ 
        12 & 1/2019 & 12/2023 & 34057 & 21 \% \\ 
        13 & 2/2019 & 12/2023 & 33007 & 22 \% \\ 
        14 & 2/2019 & 12/2023 & 33144 & 22 \% \\ 
        15 & 2/2019 & 12/2023 & 33604 & 21 \% \\ 
        16 & 2/2019 & 12/2023 & 25622 & 40 \% \\ 
        17 & 2/2019 & 12/2023 & 28414 & 33 \% \\ 
        18 & 2/2019 & 12/2023 & 25885 & 39 \% \\ 
        19 & 2/2019 & 12/2023 & 16792 & 61 \% \\ 
        20 & 2/2019 & 12/2023 & 33253 & 21 \% \\ 
        21 & 2/2019 & 9/2021 & 14238 & 37 \% \\ 
        22 & 2/2019 & 12/2023 & 26452 & 37 \% \\ 
        \hline
    \end{tabular}
    \caption{Statistics on the hourly aggregated Tethys Dataset buildings between 2019 and 2023 including hours of data expected, and percentage of missed data points.}
    \label{tab:dataset}
\end{table}

For each hour, the data we have for each building is a single, ever-increasing, number that resembles the total volume of water that has been consumed by the building since the installation of the water meter.
Due to technical issues, interference, or communication low quality in the wireless medium used to connect the water meters with the cloud infrastructure of the deployment some of the data from the infrastructure were not collected by the system. 
These missed measurements, were not transmitted redundantly due to the specifications of the system at the time, and as a result, appear as blanks in our final datataset.
These blanks can be single measurements over a week-long recording period or even whole days in some extreme cases of issues regarding the metering and/or relaying infrastructure.
From a strictly smart metering scope, missing a few data points in a week or month-long period does not create any issue, as the missing data are accumulated in the consumption of the subsequent hours. 
On the contrary, from a statistical analysis scope having blanks in a dataset requires additional analysis to accurately fill in the missing information to provide better consumption and usage profiles.

\section{Imputation Pipeline}

To achieve the best result in our methodology, we split our work into 3 steps: data preprocessing, model training, and data imputation. 
These steps will be presented in the rest of this section with operational details, and the models we evaluated.

\subsection{Data Preprocessing}
The inaugural step in our work is data pre-processing.
Our first task is to mark the missing data points of the whole dataset as a Not-A-Number value, using  Numpy \texttt{nan}.
As a second step, we split the data available in 24-hour-long vectors to benefit from the measurements' periodicity, as higher consumption is typically observed during working hours, while during nighttime the water consumed is negligible.
This 24-hour interval is used as a data entry in all further operations, and our imputation will try to fill in missing data in these vectors.
The third and final preprocessing step in our pipeline is to split the days where we have data for the whole 24-hour vector (\texttt{non-missing-days}) from the days with one or more missing data points (\texttt{missing-days}).
These \texttt{non-missing-days} are used to build our training and validation datasets, as to estimate the quality of our imputations we need to know the value we are trying to calculate.
From this data, we select 80\% for training of our models, and 20\% for validation.

\subsection{Model Training}

The next step in our pipeline is the training of the model to be used for data imputation. 
The models we evaluate in this work are the \textit{K-Nearest Neighbors} and \textit{MissForest} Imputers, a Transformer-based implementation called \textit{TimesNet}, a \textit{Multi-Directional Recurrent Neural Network} and \textit{Self-Attention-based Imputation for Time Series}.

The \textsf{K-Nearest Neighbors (KNN) Imputer} is utilized in ML to address the issue of missing data. 
This technique hinges on the concept that the missing values can be imputed by examining the nearest neighbors—those instances most akin to the one missing data. It shines when data exhibits inherent patterns or relationships, suggesting that neighboring data points can provide reasonable estimates for missing values.

On the other hand, \textsf{MissForest} stands out as a robust imputation algorithm that employs the random forest approach to tackle missing data, particularly with datasets containing both numerical and categorical variables. Its strength lies in its capacity to unearth and leverage intricate interdependencies among variables, offering an imputation precision that surpasses basic techniques like mean or median imputation.

A \textsf{Transformer} model, initially conceived for tasks in natural language processing, has been adeptly repurposed for imputation challenges, capitalizing on its proficiency with sequential data.
It deduces missing values by delving into the interplay and contextual nuances among features, harnessing the entire dataset's feature relationships through a mechanism that captures complex dependencies and fills in missing data.

\textsf{TimesNet} introduces a novel approach to transforming one-dimensional time series data into a two-dimensional matrix for each distinct period. This transformation is executed via a specialized TimesBlock module, which, through an inception-style block with shared parameters, concurrently captures the dynamics within individual periods and across multiple periods, offering a nuanced understanding of temporal patterns.

The \textsf{USGAN} is a variant of Generative Adversarial Networks (GANs) designed for unsupervised data imputation tasks. 
In data imputation, the generator learns the underlying data patterns and invents realistic values to fill in missing entries.
This is particularly useful for complex data, where simple methods like averaging wouldn't suffice.
GANs can capture relationships within the data, leading to more accurate imputations that preserve the original structure of data.

The \textsf{Multi-Directional Recurrent Neural Network (MRNN)} offers a strategic solution for imputing missing values in time series data by exploiting the temporal correlations present within the data. Utilizing Recurrent Neural Networks (RNNs), MRNN can meticulously analyze the time series' chronological patterns to predict missing values with a nuanced understanding of the temporal context.

\textsf{Self-Attention-based Imputation for Time Series (SAITS)}, represents a cutting-edge method for addressing missing values in multivariate time series data. Diverging from the MRNN's reliance on RNNs, SAITS employs a self-attention mechanism, enabling the model to selectively concentrate on pertinent segments of the time series for imputation. This mechanism meticulously assesses the interrelations across various data points and time steps, ensuring a more targeted and relevant imputation process.


For each of the algorithms above, we evaluate specific configurations that we found to work better in our use case.
This process, called a grid search, involves extensive testing of possible configurations to find what works better.
We will not go into much detail about this process, as it is a common practice and does not provide much input on the subject in question.
In more detail, we have selected the following:
\begin{itemize}
    \item \textit{KNN:} We opted for a setting of 3 neighbors.
    \item \textit{MissForest:} $4$ estimators, maximum depth of $10$, enabled bootstrap sampling, restricted max samples to $50\%$ of the data, and utilized $2$ parallel jobs for computation.
    \item \textit{SAITS:} We implemented a configuration comprising two transformer encoder layers, with a model dimensionality of $256$.
Within each transformer layer, the feed-forward network operated at a dimensionality of 128.
We employed a multi-head attention mechanism with $4$ attention heads, utilizing key and value vectors of dimensionality $64$ in the attention mechanism. 
To prevent overfitting, a dropout rate of $0.1$ was applied globally, while the attention layers specifically utilized a dropout rate of $0.1$.
    \item \textit{Transformer:} The neural network comprises $6$ transformer layers, with a model dimensionality of $256$. Within each transformer layer, the feed-forward network operated with a dimensionality of $256$. 
The multi-head attention mechanism employed $4$ attention heads, with key and value vectors of dimensionality $128$. 
To mitigate overfitting, a global dropout rate of $0.1$ was applied, while the attention layers utilized a dropout rate of $0$.
    \item \textit{Timesnet:} The neural network consists of a single TimesNet block. We employed a top-k pooling mechanism with a k-value of $1$. The model dimensionality was set to $128$, with a feed-forward network dimensionality of $512$. Utilizing $5$ kernels, the network operates with a dropout rate of $0.5$ to prevent overfitting. 
    \item \textit{USGAN:} RNN with a hidden size of 256, leveraging adversarial training to generate realistic data points for missing values. Employing an MSE loss term with a weight of 1, with dropout regularization of 0.1 to prevent overfitting. During training, the generator and discriminator networks are updated once per step.

    \item \textit{MRNN:} $3$ layer model with a hidden layer size of $256$. 
    \item All the neural network models utilize the \textit{Adam} optimizer with a learning rate of $0.001$. Each model was trained for $100$ epochs, with a model-saving strategy of \textit{best}, ensuring that the best-performing model is selected.
\end{itemize}
 

\subsection{Data Imputation}
The imputation is the final step of our pipeline and refers to the calculation of missing values using the trained models.
Data imputation in our application is done over 24-hour long periods.
To impute a single missing data point, we need to provide our models with a 24-hour values vector, one or more of which are missing.
The models used are capable of calculating more than one missing value in each vector at the same time, to recover the missing data more effectively.
For this evaluation, we focus on imputing a single blank value in each vector but the process can be extended to estimate more. 

\section{Evaluation}
For the evaluation of our work, which we showcase in the rest of this section, we \textit{simulate} missing values by hiding data points in $20\%$ of our total dataset. 
We then impute the values and compare our estimates with the actual data in the dataset.
We also evaluate all selected algorithms in two settings.
First, we run the training and evaluation using dedicated models for each building, to take advantage of the specific characteristics each building has, and its water consumption patterns.
In a second run, we build a single model with combined water consumption data and evaluate it to understand its behaviour.

For our evaluation, we use the Mean Absolute Error (MAE) metric to estimate the estimations' accuracy. It is a commonly used metric in ML to quantify the accuracy of numerical predictions measuring the average magnitude of errors in a set of predictions, without considering the direction of the errors.



\subsection{Using dedicated models}
Starting with dedicated models, we select 80\% of the dataset we have as the training dataset.
This dataset will be the same for training each model, to ensure that each algorithm is trained over the same data.
Then, in the 20\% of the data we selected as a validation dataset, we randomly replaced one measurement with a not-a-number value indicating it is missing.
Again, the replaced values are the same across all models, to ensure the fairness of our evaluation.

\begin{figure}[h]
    \centering
    \includegraphics[width=.45\textwidth]{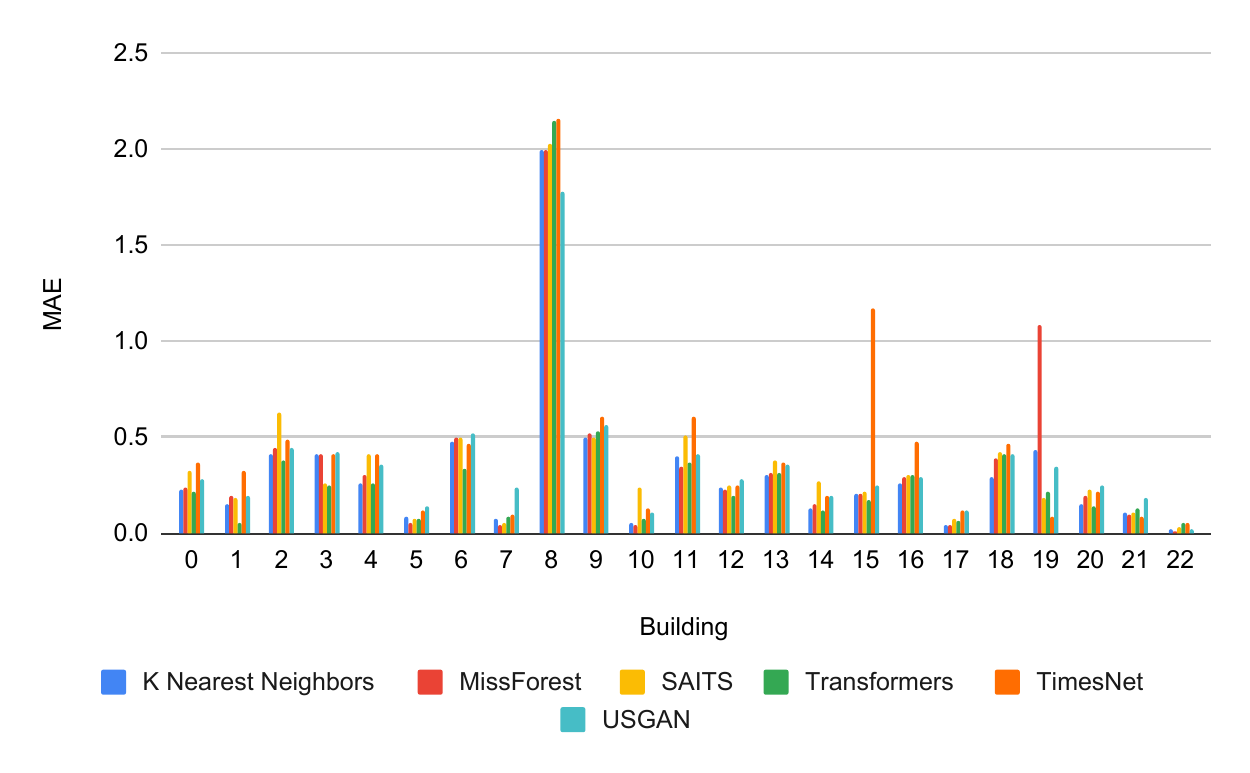}
    \caption{MAE for each dedicated model for all 22 buildings of our dataset.}
    \label{fig:stats-for-buildings}
\end{figure}

Figure~\ref{fig:stats-for-buildings} presents the MAE for each model in each building's dataset, showcasing how each method provides better results in some buildings while not performing so well in others.
This behavior is attributed to the actual values of the data available in for building as in some cases, some consumption patterns are hard to predict when focusing on a single building, like out-of-profile, irregular, daily consumption patterns.
In general, the Transformer model and the kNN model perform best in this setting.
We also observe that in building 8 all algorithms perform badly. 
This is attributed to the very high amount of data missing from this specific building with over 79\% of the data lost during the testbed's lifetime.

\begin{figure}[h]
    \centering
    \includegraphics[width=.45\textwidth]{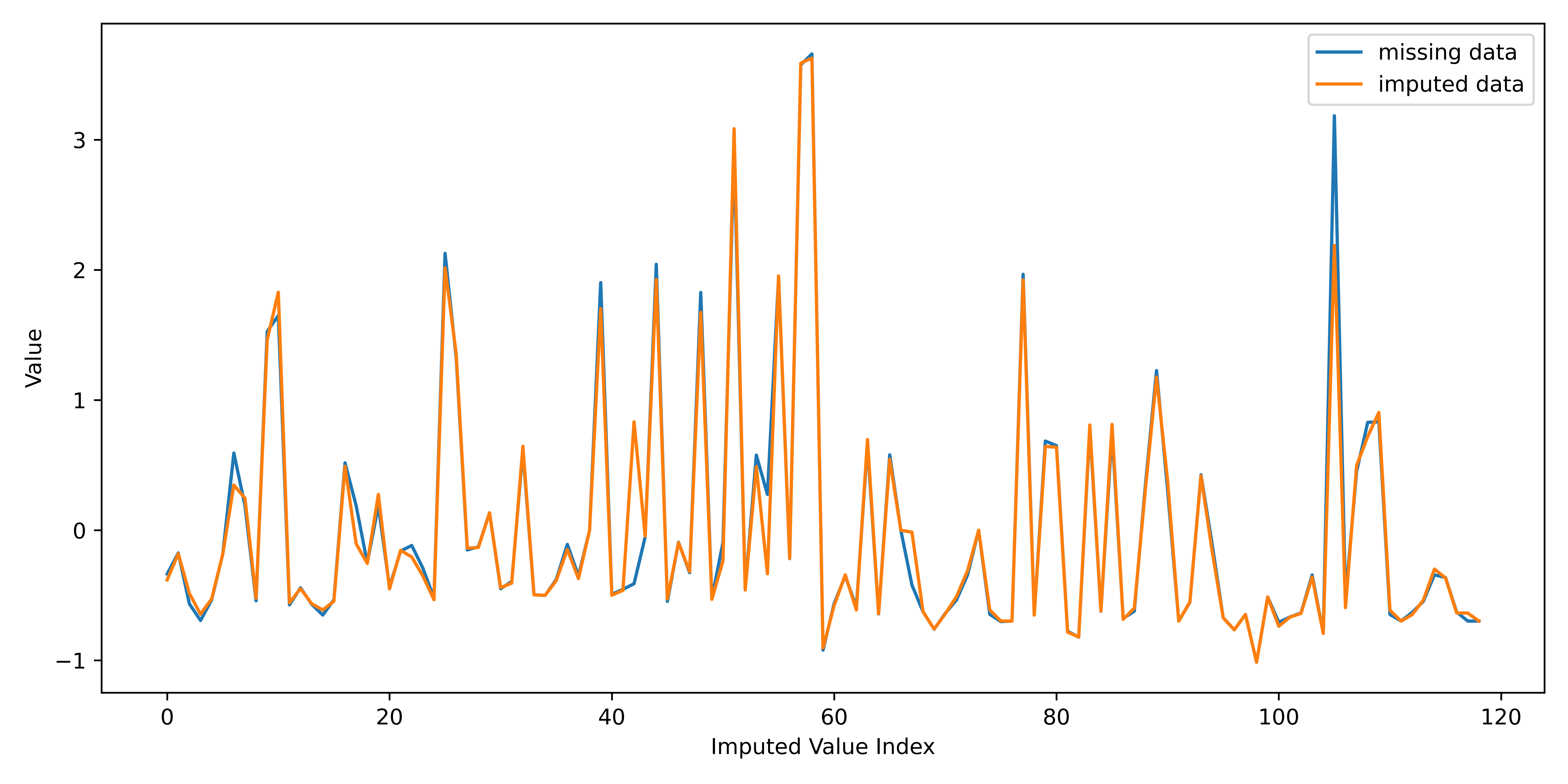}
    \caption{Actual \& Imputed data using the dedicated kNN model (building 10).}
    \label{fig:knn-12}
\end{figure}

\begin{figure}[h]
    \centering
    \includegraphics[width=.45\textwidth]{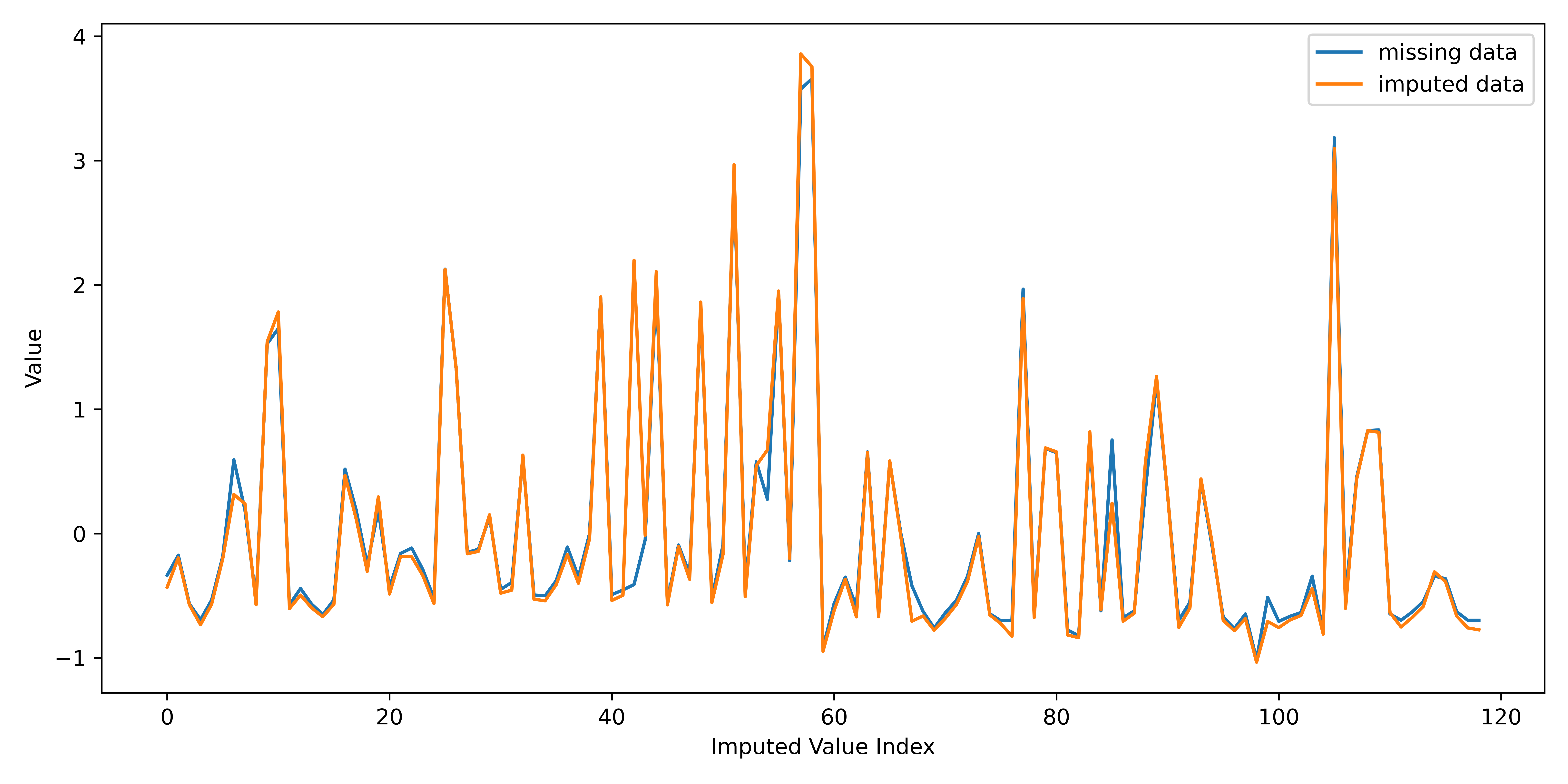}
    \caption{Actual \& Imputed data using the dedicated Transformer model (building 10).}
    \label{fig:msf-8}
\end{figure}

Diving into more detail for the two best-performing models, Figure~\ref{fig:knn-12} showcases the comparison of the imputed and actual values for building 10 using the kNN model, while Figure~\ref{fig:msf-8} depicts the performance of the Transformer model for building 10 providing better estimates for the same missing values. 

\subsection{Using common models}
Moving on to the common models, we again select 80\% of the merged dataset we have as the training dataset.
This dataset is the same for training each model, to ensure that each algorithm is trained over the same data.
Then, in the 20\% of the data we selected as a validation dataset, we randomly replaced one measurement with a not-a-number value indicating it is missing. Again, the replaced values are the same across all models, to ensure evaluation fairness.

\begin{figure}[h]
    \centering
    \includegraphics[width=.45\textwidth]{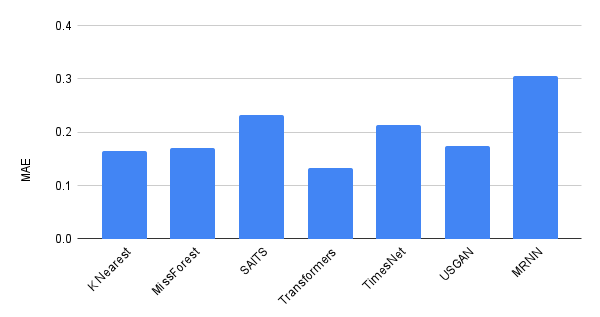}
    \caption{MAE for the common model for all 22 buildings of our dataset.}
\label{fig:stats-for-buildings-common}
\end{figure}

\begin{figure}[h]
    \centering
    \includegraphics[width=.45\textwidth]{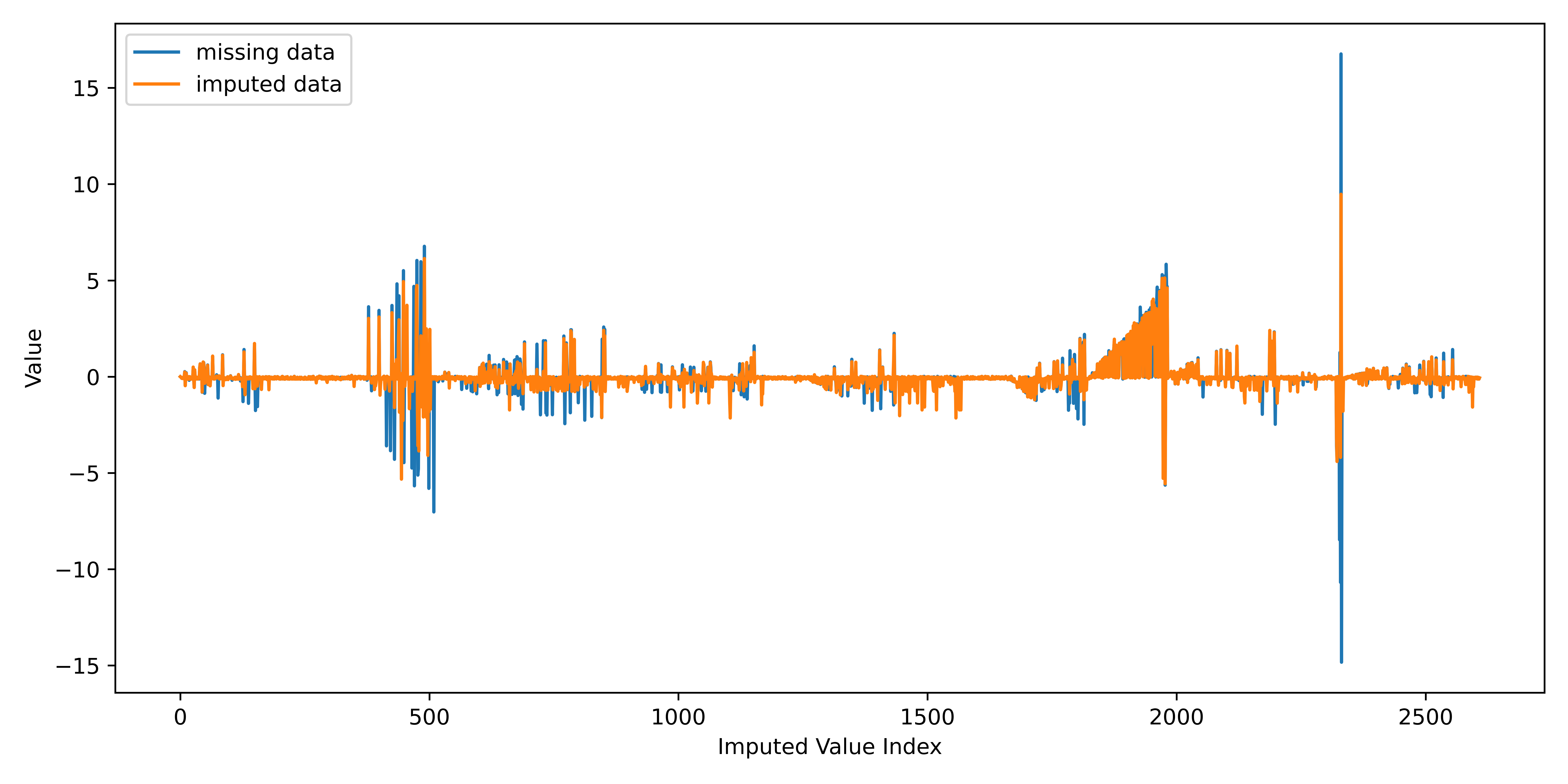}
    \caption{Actual and Imputed values using the common kNN model.}
    \label{fig:c-msf-8}
\end{figure}

Figure~\ref{fig:stats-for-buildings-common} showcases the MAE for each model in each building's dataset.
As we can observe from the figure, some models again provide us with better results, like the Transformer model, kNN and MissForest, that did not perform so well in the previous case.
The worst performing model is once again MRNN, which indicates that some more investigation on the model's hyperparameters is needed.
Figure~\ref{fig:c-msf-8} showcases the performance of kNN in estimating all the missing data from all the buildings, achieving the best performance in both settings.

\subsection{Comparison of the dedicated and common models}

Table~\ref{tab:averages} contains the average MAE values for each evaluated model, both in the dedicated and common model settings.
We can see that increasing the dataset helped to generate better estimates in total, allowing for more accurate predictions.
As noted before, this is attributed to performing on average better in all cases, including out-of-profile consumption patterns that may exist in each building that are better covered in the common model using data from other buildings.

\begin{table}[h]
    \centering
    \begin{tabular}{|c|c|c|}
    \hline
        Model & MAE dedicated & MAE common \\
    \hline
        kNN & 0.31 & 0.16 \\
        MissForest & 0.35 & 0.17 \\
        SAITS & 0.35 & 0.23\\
        Transformer & 0.30 & 0.13\\
        TimesNet & 0.42 & 0.21\\
        USGAN & 0.35 & 0.17\\
        MRNN & 0.62 & 0.30\\
    \hline        
    \end{tabular}
    \caption{Average MAE values for each dedicated model type compared to the common models.}
    \label{tab:averages}
\end{table}

\section{Conclusions}
In our work, we studied the application of multiple data imputation techniques to address challenges associated with missing data in water grid monitoring systems equipped with smart IoT meters.
Through meticulous comparison and evaluation of different methods, including K-Nearest Neighbors, MissForest, and neural network-based approaches like SAITS and TimesNet, we identified optimal strategies tailored to the specific needs of water distribution systems. 
The integration of such imputation techniques in monitoring systems resulted in filling in the blanks in the original dataset with low deviations from the actual values and a mean absolute error as low as 0.16.
This enhancement is crucial for effective decision-making in water conservation, leak detection, and infrastructure maintenance, ultimately leading to increased operational efficiency and sustainability in water management practices.
Next steps in our path include the use of the proposed system in the metering environment, as a real-time application that can predict and fill blanks, available to each customer, to collect feedback on the quality and accuracy of the predictions.



\bibliographystyle{IEEEtran}
\bibliography{conference_101719}

\end{document}